%% file: main.tex
\documentclass[runningheads,oneside]{llncs}

\input{preamble_arxiv}

\begin{document}

\input{sec/00_teaser}

\input{sec/01_abstract}

\input{sec/02_intro}
\input{sec/03_rel_works}
\input{sec/04_prelims}

\input{sec/05_method}
\input{sec/06_experiments}
\input{sec/07_conclusion}
{
    \small
    \bibliographystyle{splncs04}
    \bibliography{main}
}

\end{document}

%% file: preamble_arxiv.tex
\usepackage[final,year=2026,ID=4399]{eccv}
\usepackage{eccv}

\usepackage{eccvabbrv}
\usepackage[pagebackref,breaklinks,colorlinks,allcolors=eccvblue]{hyperref}
\usepackage{graphicx}
\usepackage{booktabs}
\usepackage[accsupp]{axessibility}  
\usepackage{orcidlink}
\usepackage{multirow}
\usepackage{booktabs}
\usepackage{xcolor}
\usepackage{algorithm}
\usepackage[noend]{algpseudocode}
\usepackage{pifont}
\usepackage{capt-of}
\usepackage{csquotes}
\usepackage{balance}
\usepackage[table]{xcolor}
\usepackage{makecell}

\usepackage{graphicx}
\usepackage{amsmath,amssymb}
\usepackage{booktabs}

\authorrunning{Bose et al.}
\titlerunning{\ours{}}

\newcommand{\cmark}{\textcolor[HTML]{009901}{\ding{51}}} 
\newcommand{\xmark}{\textcolor{red}{\ding{55}}}          

\definecolor{pink}{HTML}{FF99FF}
\definecolor{brown}{HTML}{6D4C41}

\newcommand{\red}[1]{{\color{red}#1}}

\newcommand{\cyan}[1]{\textcolor{cyan}{#1}}
\newcommand{\green}[1]{\textcolor{green}{#1}}
\renewcommand{\green}[1]{\textcolor[HTML]{009901}{#1}}

\algrenewcommand{\algorithmiccomment}[1]{%
  \hfill\textcolor{cvprblue}{\scriptsize #1}%
}

\definecolor{oursrow}{HTML}{F3F7FF} 

\definecolor{baselinebg}{HTML}{FFF3E6}  
\definecolor{advancebg}{HTML}{EAF7FF}   
\definecolor{blockhdr}{HTML}{FFFFFF}    

\newcommand{\ours}[1]{\textcolor{blue}{\texttt{VALOR}}}

\definecolor{cvprblue}{rgb}{0.21,0.49,0.74}

\title{Visual Alignment of Medical Vision-Language Models for Grounded Radiology Report Generation}

\author{
Sarosij Bose\inst{1,2}\thanks{$^\dagger$ Work done during internship at NEC Laboratories America} \and
Ravi K. Rajendran\inst{1} \and
Biplob Debnath\inst{1} \and
Konstantinos Karydis\inst{2} \and
Amit K. Roy-Chowdhury\inst{2} \and
Srimat T. Chakradhar\inst{1}
}

\institute{
NEC Laboratories America \and
University of California, Riverside
}

%% file: sec/00_teaser.tex
\maketitle

\begin{center}
\vspace{-1em}
\begin{minipage}{\textwidth}
  \centering
  \includegraphics[width=\linewidth, height=0.4\textheight, keepaspectratio]{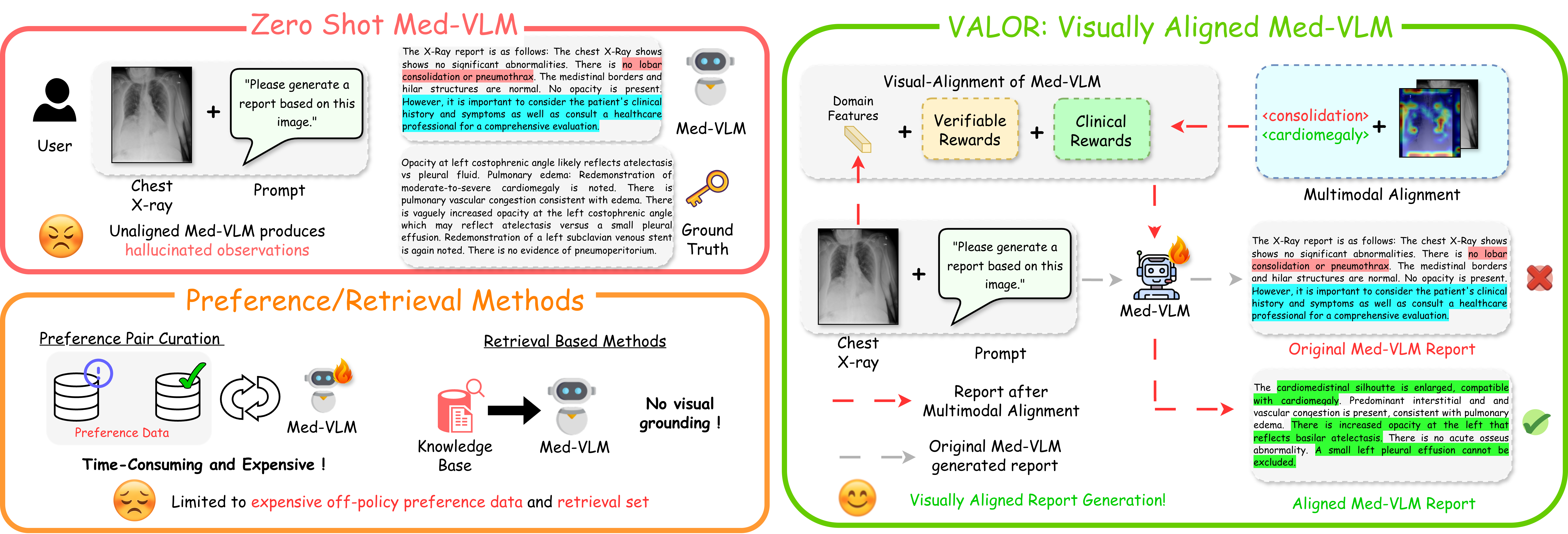}

  \captionof{figure}{%
    \textbf{Overview:} We introduce \ours{}, a multimodal alignment framework that addresses the visual hallucinations of existing Med-VLMs~\cite{li2023llava}. 
    \textbf{Left:} the generated response from vanilla Med-VLM where the misalignment with the input image is marked in \red{red} and generic non-medical text in \cyan{cyan}. 
    Existing approaches resort to curation of preference data per-task or on retrieval of reports which is inherently expensive and does not address the issue of multimodal alignment. 
    \textbf{Right:} Our method \ours{}, addresses this issue with a novel visual reasoning based pipeline which utilizes multi-modal rewards for optimization and pushes the model to generate more clinically informed and grounded reports aligned with the original image which are marked in \green{green}.%
  }
  \label{fig:teaser}
\end{minipage}
\vspace{1em}
\end{center}

%% file: sec/01_abstract.tex
\begin{abstract}
Radiology Report Generation (RRG) is a critical step toward automating healthcare workflows, facilitating accurate patient assessments, and reducing the workload of medical professionals. Despite recent progress in Large Medical Vision-Language Models (Med-VLMs), generating radiology reports that are both visually grounded and clinically accurate remains a significant challenge. Existing approaches often rely on large labeled corpora for pre-training, costly task-specific preference data, or retrieval-based knowledge. However, these strategies do not adequately mitigate hallucinations arising from poor cross-modal alignment between visual and linguistic representations. To address these limitations, we propose \textbf{\ours{}}: \underline{V}isual \underline{A}lignment of Medical Vision-\underline{L}anguage Models for Gr\underline{O}unded \underline{R}adiology Report Generation, which tackles visual hallucinations through two complementary reasoning stages: (1) Clinically Informed Textual Reasoning guides the model with verifiable natural language and clinical metric rewards to produce semantically complete reports with precise medical terminology. (2) Self-Supervised Visual Reasoning leverages a frozen domain expert to compute image-text similarity scores between the input chest X-ray and generated candidates, converting these into rank-normalized advantages that explicitly steer the policy toward visually grounded outputs, requiring no preference pairs, retrieval databases, or additional annotations. Extensive experiments on multiple benchmarks demonstrate that \ours{} substantially improves generation quality, as well as clinical accuracy which are visually grounded, achieving significant performance gains over state-of-the-art medical report generation benchmarks.
\keywords{Report Generation \and Visual Alignment and Reasoning}
\vspace{-0.5em}
\end{abstract}

%% file: sec/02_intro.tex
\section{Introduction}\label{sec:introduction}
Automated chest X-ray report generation has emerged as a promising avenue to reduce the burden of medical professionals for manual radiology reporting. This has emerged with the rise of Large Medical Vision-Language Models (Med-VLMs)~\cite{li2023llava, zhou2023skingpt, zhou2023path, sellergren2025medgemma, nath2025vila} becoming the dominant paradigm, leveraging large-scale multimodal pretraining and external clinical knowledge to generate reports directly from images. However, these models frequently produce reports that are not grounded in the input image, generating clinically plausible but unsupported findings, a phenomenon known as ``Visual Hallucination'' ({\hyperref[fig:teaser]{Fig.~\ref{fig:teaser}}}). This arises primarily because Med-VLMs tend to rely on learned patterns arising from text-heavy pretraining corpora rather than performing genuine cross-modal reasoning~\cite{leng2024mitigating, Huang_2024_CVPR, liu2025gemex, amirloo2024understanding, cao2024madtp, zhou2024aligning}. The problem is further compounded in retrieval-augmented pipelines, where clinical statements from retrieved reports can be incorporated even when they are inconsistent with the visual evidence~\cite{xia2024rule, hou-etal-2025-radar, chen2024detecting}. This prompts a fundamental question: \textit{\textbf{Can we design a framework that generates clinically informative reports which are both factually accurate and visually grounded in the image?}}

Existing efforts to mitigate visual hallucinations face two fundamental limitations: (1) \textbf{Data Reliance:} Methods that rely on preference datasets~\cite{chen2024huatuogpt, cui2024biomedical, ouyang2022training, zhu2024mmedpo} are inherently \textit{bounded by the quality of their preference pairs}, which are typically constructed using simplistic techniques devoid of clinical context. More critically, the preference optimization usually happens entirely in the textual space: the model learns to prefer one report over another without any mechanism to verify whether either report is actually supported by the input image and does not account for the inherent gap between reports and provided images (2) \textbf{Visual Misalignment:} Retrieval-augmented approaches~\cite{jin2021disease, xia2024rule, sun2025fact, hou-etal-2025-radar} inject external clinical context, but introduces a new source of hallucination, since \textit{retrieved reports describe other patients' images and may contain findings entirely absent from the current X-ray} ({\hyperref[fig:teaser]{Fig.~\ref{fig:teaser}}}). Further, this requires training retrieval models on domain specific data which adds computational overhead. The semantic gap between retrieved text and the visual evidence is fundamental, not merely a matter of retrieval quality. As a result, neither paradigm enforces what is most needed: \textbf{\textit{direct alignment between the generated text and the visual content of the image, without relying on external preference data, retrieval databases, or additional annotations}} ({\hyperref[fig:sota_chart]{Fig.~\ref{fig:sota_chart}}}).


\input{figures/sota_radar_chart}

To address these challenges, we propose \textbf{\ours{}}: \emph{\underline{V}isual \underline{A}lignment of Medical Vision-\underline{L}anguage Models for Gr\underline{O}unded \underline{R}adiology Report Generation}. \ours{} enhances the medical reasoning of Med-VLMs through a two-stage multimodal optimization process. In the first stage, \textbf{Clinically Informed Textual Reasoning}, the model is guided by verifiable natural language generation metrics with \textit{explicit clinical guidance} as rewards to produce semantically coherent reports with precise clinical terminology. In the second stage, \textbf{Self-Supervised Visual Reasoning}, we leverage a frozen multi-label domain expert model to compute image-text similarity scores in-between the input X-ray and each generated candidate reports per rollout, converting these into \textbf{rank-normalized advantages} that explicitly steer the policy toward visually grounded outputs. In summary, our main contributions are as follows:

\begin{enumerate}
    \item We propose \textbf{\ours{}}, a novel framework that generates \textbf{visually grounded and clinically informative} radiology reports through two complementary reasoning stages, reducing visual hallucinations and strengthening image-report consistency without relying on any external curated preference datasets or domain specific retrieval-based methods.

    \item We introduce \textbf{clinically-informed textual rewards} for precise terminology in the generated report and \textbf{rank-normalized advantages}, which utilizes image-text similarity scores obtained from a frozen domain expert for visual guidance, allowing the policy to distinguish which candidate reports are closer to the input X-ray. This enables visual grounding \textbf{without requiring preference pairs, retrieval databases, or additional annotations}, using only the agreement between the generated report and disease-relevant features for grounded report generation.  


    \item We conduct extensive evaluations of \ours{} on IU-XRay~\cite{demner2015preparing} and MIMIC-CXR~\cite{johnson2020mimic}, reporting both generation quality and clinical accuracy metrics, which are rarely assessed jointly in prior work. We also provide a qualitative visual grounding analysis through attention visualizations. As shown in {\hyperref[fig:sota_chart]{Fig.~\ref{fig:sota_chart}}}, \ours{} consistently outperforms both baseline and state-of-the-art preference-alignment methods by approximately 20\% in generation quality and 25\% in clinical accuracy. Furthermore, in a zero-shot setting, \ours{} surpasses medical and proprietary Med-VLMs by approximately 17\%, generating more coherent and visually grounded reports that focus on clinically meaningful regions with improved factual accuracy.
\end{enumerate}

%% file: figures/sota_radar_chart.tex
\begin{figure}[t]
    \centering
    \begin{minipage}[c]{0.5\linewidth}
        \centering
        \includegraphics[width=\linewidth]{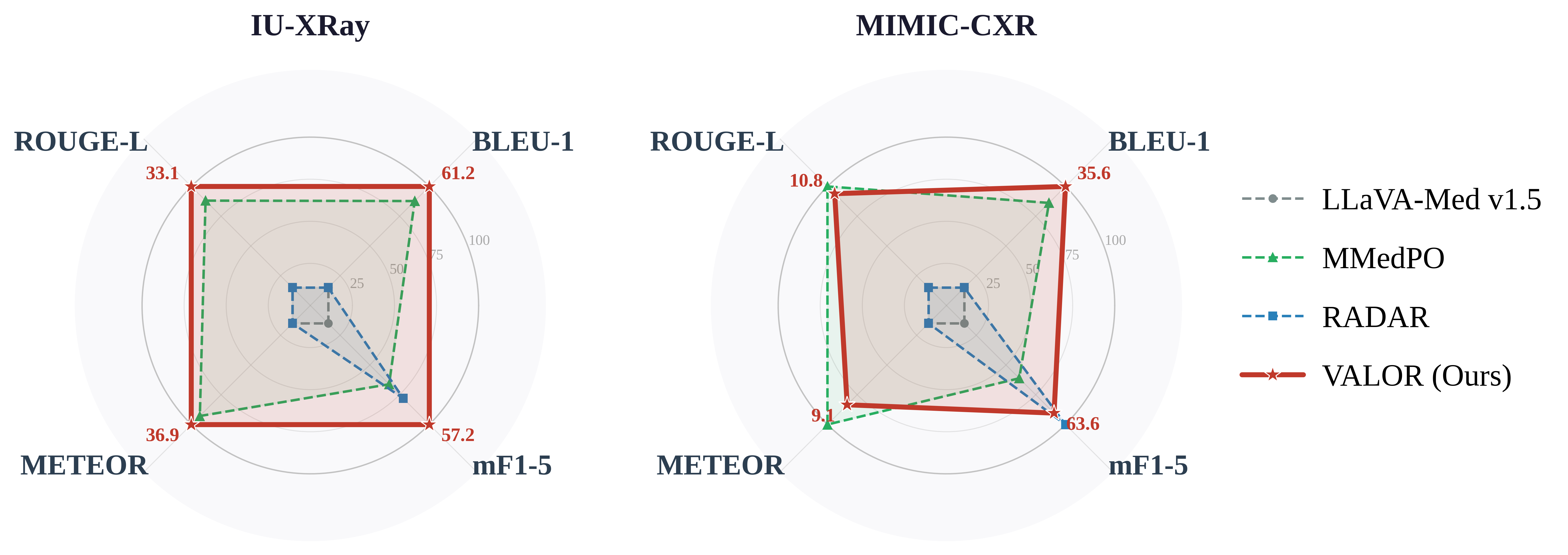}
    \end{minipage}%
    \hfill
    \begin{minipage}[c]{0.45\linewidth}
        \centering
        \setlength{\tabcolsep}{3pt}
        \resizebox{\linewidth}{!}{%
        \begin{tabular}{lccc}
        \toprule
        \textbf{Model} & \makecell{\textbf{No Pref.}\\\textbf{Data}} & \textbf{No Retrieval} & \makecell{\textbf{Visual}\\\textbf{Grounding}} \\
        \midrule
        LLaVA-Med v1.5~\cite{li2023llava} & -- & -- & \xmark \\
        MMedPO~\cite{zhu2024mmedpo}         & \xmark & \cmark & \xmark \\
        RADAR~\cite{hou-etal-2025-radar}          & \cmark & \xmark & \xmark \\
        \rowcolor{oursrow}
        \textbf{\ours{}} & \cmark & \cmark & \cmark \\
        \bottomrule
        \end{tabular}%
        }
    \end{minipage}
    \caption{\textbf{Comparison with SOTA methods.} \ours{} performs significantly better in generation quality~\cite{papineni2002bleu, lin2004rouge, denkowski2011meteor} and clinical accuracy~\cite{irvin2019chexpert, ostmeier2024green, zhao2024ratescore} without using any preference-data or additional retrieval-based methods. Further, these works do not perform disease-specific visual grounding w.r.t. the given image as shown in the table.}
    \label{fig:sota_chart}
    \vspace{-1em}
\end{figure}

%% file: sec/03_rel_works.tex
\section{Related Works}\label{sec:related}

\textbf{Radiology Report Generation (RRG).} Radiology report generation involves composing descriptive sentences that capture the contents of a medical image, organized into a ``Findings'' section for detailed observations and an ``Impression'' section for fine-grained interpretation that is clinically informed and aligned with the image. Various works have been proposed along these lines~\cite{huang2023kiut, jin2024promptmrg, li2023dynamic}.
Existing approaches for RRG can be broadly categorized into three groups. The first focuses on directly improving the encoder-decoder architecture to generate more consistent and medically enriched reports. Earlier approaches used LSTM~\cite{hochreiter1997long} based networks with hierarchical structures to manage the descriptiveness of radiology reports~\cite{tanida2023interactive}, dynamic graphs with a contrastive learning paradigm~\cite{li2023dynamic} or supervised multimodal pre-training~\cite{tanida2023interactive, wang2023metransformer, park2025dart}. The second group of approaches focuses on closest report retrieval using retrieval-augmented generation (RAG) techniques, where a retriever is trained based on medical informativeness and targeted retrieval using domain knowledge~\cite{li2024rag, dong2025understand, xia2024mmed, chu2025reducing}. RADAR~\cite{hou-etal-2025-radar} employs the combination of retrieved reports and the overlap with the potential diseases which may be there in the image. However, these methods either employ paired multimodal data, require additional retrieved information during training or inference and generate terminology that is misaligned with the image, limiting their reliability in clinical practice.

\noindent\textbf{Reinforcement Learning in LLMs/VLMs.} Despite the rise of Supervised Fine-Tuning (SFT)-based techniques, Reinforcement Learning (RL)-based approaches have also gained popularity owing to their robustness, scalability, ability to align with human preferences, and ability to mitigate hallucinations~\cite{yu2025rlaif, zhou2024aligning, wang2025instruction}. MMedPO~\cite{zhu2024mmedpo} utilizes a Direct Preference Optimization (DPO)-based approach based on curating preference data to mitigate hallucinations in Med-VLMs. However, it requires the curation of preference data, which is time-consuming and expensive, and does not generalize well outside of the curated preference data. Our work focuses on addressing this limitation by proposing a novel two-stage post-training recipe that
focus on unifying both modalities, and progressively enhances the visual and clinical reasoning capabilities of the model.

\noindent\textbf{Medical Reasoning using Vision Language Models.} With the widespread success of CLIP~\cite{radford2021learning}, several other VLMs have arisen in the medical space such as Med-Flamingo~\cite{moor2023med}, BioMedGPT~\cite{zhang2024generalist} and HuatuoGPT-Vision~\cite{chen2024huatuogpt}which learn the cross-modal information across a large scale of pre-training data across various vision and language modalities such as CheXzero~\cite{tiu2022expert}, Med-CLIP~\cite{wang2022medclip} and BioMedCLIP~\cite{zhang2023biomedclip} which focus on enhancing the understanding of raw reports with images whereas BioVIL~\cite{boecking2022making}, BioVIL-T~\cite{bannur2023learning} enhance the encoding of the textual descriptions. More recent work, such as MedKLIP~\cite{wu2023medklip} leverages the spatial relationships from RadGraph~\cite{jain2021radgraph} to incorporate more fine-grained medical information into the pre-training regimen. Methods such as MMedAgent~\cite{li2024mmedagent} and VILA-M3~\cite{nath2025vila} utilize expert knowledge (requiring annotation) and tool usage (prone to model noise) as additional proxies of information for alignment to the medical domain. Thus, these successes can be largely attributed to the availability of a large corpus of labeled pretrained data and largely ignore the alignment of human preferences with the provided image. It is therefore important to move away from external dependencies such as tools, preference data or retrieval datasets and adjust the inherent reasoning focus of the model towards the provided X-ray. \emph{Our novel post-alignment framework aims to address this open challenge of improving the visual reasoning abilities of Med-VLMs without using additional external knowledge in a self-supervised fashion.}


%% file: sec/04_prelims.tex
\section{Preliminaries}
\label{sec:prelim}

\textbf{Problem Formulation.}
Given a 2D chest X-ray image $\mathcal{I} \in \mathbb{R}^{H \times W \times 3}$, a text prompt $\mathcal{P}$, and a dataset $\mathcal{D} = \{(\mathcal{I}, \mathcal{R}_{\text{gt}})\}$ of images and ground truth reports, the goal of radiology report generation is to produce a report $\mathcal{R}$ that is clinically accurate and aligned with the input image $\mathcal{I}$. We model this with a base Med-VLM $\mathcal{M}_\theta$, which induces a conditional policy over it's parameters $\pi_\theta(\mathcal{R} \mid \mathcal{I})$ to generate $\mathcal{R}$ in an autoregressive manner. In this work, we adapt $\mathcal{M}_\theta$ in two stages: first, we refine the text generation policy using verifiable and clinical textual rewards; second, we further align the model with image-conditioned visual rewards derived from a frozen domain expert $\mathcal{G}$, encouraging reports whose disease findings are grounded in the underlying X-ray $\mathcal{I}$.

\noindent\textbf{RRG Objective.} We follow existing works and process reports in two parts: the \texttt{FINDINGS} and \texttt{IMPRESSION} sections, where the former contains the broad rationale behind the generated report and the latter contains the potential diseases present in the patient's medical image. Formally, this process can be represented as follows: given the medical image $\mathcal{I}$, the Med-VLM $\mathcal{M}(\cdot)$ is tasked with interpreting the image, reasoning about it, and generating a descriptive report $\mathcal{R} = \{ r_1, r_2, \ldots, r_T \}$ where $T$ represents the total length of the report and $r_{t} \in \mathbb{V}$ represents the tokens in the vocabulary. The report $\mathcal{R}$ is auto-regressively generated as,
\begin{equation}
p(\mathcal{R} \mid \mathcal{I}) = \prod_{t=1}^{T} p(r_t \mid r_1, \ldots, r_{t-1}, \mathcal{I}).
\end{equation}

%% file: sec/05_method.tex
\input{figures/pipeline}
\section{Methodology}\label{sec:method}
We describe the \ours{} framework below in two parts: (1) \textbf{Instruction-Following Warmup}, which stabilizes the Med-VLM for structured report generation and prevents degenerate outputs during subsequent optimization, and (2) \textbf{Multi-Stage Rewards for Clinical and Visual Alignment}, which progressively guides the policy toward clinically precise and visually grounded reports.

\subsection{Instruction-Following Warmup}\label{subsec:warmup}
We perform a supervised fine-tuning (SFT) phase to enhance the instruction following capabilities of the Med-VLM $\mathcal{M}(\cdot)$ for radiology report generation. SFT is carried out on curated data to improve formatted generation and stabilize subsequent RL-based optimization~\cite{sun2024aligning, zhu2024mmedpo}. Given the SFT dataset $\mathcal{D}_\text{sft} = \{(\mathcal{I}, \mathcal{R}_{\text{sft}})\}$ of chest X-rays $\mathcal{I}$, prompts $\mathcal{P}$, and reports $\mathcal{R}_{\text{sft}}$, we update the model parameters of $\mathcal{M}(\cdot)$ by minimizing the negative log-likelihood w.r.t. ground-truth tokens,
\begin{equation}
\label{eq:sft}
\resizebox{0.6\columnwidth}{!}{$
\mathcal{L}_{\mathrm{SFT}}(\theta)
=
\mathbb{E}_{(\mathcal{I}, \mathcal{R}_{\text{sft}}) \sim \mathcal{D}}
\Bigg[
  - \sum_{t=1}^{T}
  \log \pi_{\theta}\big(r^{\text{sft}}_{t} \,\big|\, \mathcal{I}, r^{\text{sft}}_{<t}\big)
\Bigg]
$}
\end{equation}
where $\mathcal{R}_{\text{sft}} = (r^{\text{sft}}_{1}, \dots, r^{\text{sft}}_{T})$ denotes the tokenized report, and $\pi_{\theta}$ is the conditional distribution induced by $\mathcal{M}(\cdot)$. This SFT warmup teaches the model to respect the report structure and produce structurally well-formed reports conditioned on the input image $\mathcal{I}$ and prompt $\mathcal{P}$. In practice, we find that starting from an SFT checkpoint leads to faster convergence and avoids degenerate ``cold-start'' tokens. This SFT fine-tuned policy $\pi_{\theta}$ is then further trained using our multistage reward policy described below.

\subsection{Multi-Stage Rewards for Clinical and Visual Alignment}\label{subsec:rewards}
To guide the policy toward radiology knowledge and visually grounded outputs, we design a multi-tier reward system. The first stage focuses on semantic alignment, format adherence, and clinical accuracy using textual rewards, whereas the second phase emphasizes image–text alignment between the generated report and input image, injecting disease-relevant visual reasoning.

\input{algo/algo}

\noindent\textbf{\underline{Clinically Informed Textual Reasoning.}}
Here, we adapt the Med-VLM using \emph{verifiable} text rewards that can be computed directly from the ground truth report. We combine NLG Metrics~\cite{papineni2002bleu, lin2004rouge}, which emphasize report token similarity, with completeness~\cite{zhang2019bertscore}, which measures semantic coherence while being less sensitive to generic template text. At each alignment step, $\mathcal{M}(\cdot)$ rolls out a set of candidate reports $\{ o_i \}_{i=1}^{G}$ from the current policy $\pi_{\theta}$. For each candidate report $o_i$ and ground truth report $g$, the verifiable text reward is given as,

\begin{equation}
\resizebox{0.9\columnwidth}{!}{$
\begin{aligned}
R_{\text{ver}}(o_{i},g)
&= \lambda_{\text{lex}}
   \big[\mathrm{BLEU}_4(o_{i},g)+\mathrm{ROUGE}_L(o_{i},g)] + \lambda_{\text{sem}}\,\mathrm{BERTScore}(o_{i},g),
\end{aligned}
$}
\label{eq:ver-reward}
\end{equation}
where weight hyperparameters $\lambda_{\text{lex}},\lambda_{\text{sem}}\!\ge 0$ and $\lambda_{\text{lex}}+\lambda_{\text{sem}}=1$.

To encourage clinically correct disease terms, we use the micro-averaged F1 scores from CheXbert and RadGraph entities as \emph{periodic} rewards.  
Let $s_{\text{cx}},s_{\text{rg}}\!\in[0,1]$ denote normalized F1 scores between labels (or triplets) and entities extracted from $o_i$ and $g$.
We define the clinical reward as,
\begin{equation}
\resizebox{0.5\columnwidth}{!}{$
R_{\text{clin}}(o_{i},g)
= \lambda_{\text{cx}}\,s_{\text{cx}}(o_{i},g)
+ \lambda_{\text{rg}}\,s_{\text{rg}}(o_{i},g),
$}
\label{eq:clin-reward}
\end{equation}
with $\lambda_{\text{cx}},\lambda_{\text{rg}}\!\ge 0$ as weighting hyperparameters.

Rather than applying clinical rewards at every policy step, which destabilizes the optimization, we inject them only every $k_{\text{clin}}$ updates. Let $k$ denote each rollout from the global policy update step.  
Then, the Stage-1 scalar reward used to compute group relative advantages is combined as,
\begin{equation}
\resizebox{0.75\columnwidth}{!}{$
R_{\text{text}}(o_{i},g;k)
=
\begin{cases}
R_{\text{ver}}(o_{i},g) + \lambda_{\text{clin}}\,R_{\text{clin}}(o_{i},g), &  \text{if } k = k_{\text{clin}},\\[4pt]
R_{\text{ver}}(o_{i},g), & \text{otherwise},
\end{cases}
$}
\label{eq:stage1-text}
\end{equation}
where $\lambda_{\text{clin}}\!\ge 0$ controls the strength of the periodic clinical contribution.
This design provides dense, verifiable supervision at every step while periodically nudging the policy toward clinically faithful language without allowing natural language metrics to entirely dominate the training process.

\noindent\underline{\textbf{Self-Supervised Visual Reasoning.}}
In Stage-2, we introduce image-text similarity scores that explicitly tie each candidate report with the provided X-Ray. 
For an image $I$, we obtain a continuous domain feature vector $\mathcal{F}_D = \mathcal{G}(\mathcal{I})$ from the frozen expert model, and for each sampled candidate $o_i$, we compute a text embedding $\mathbf{e}_i^{\text{txt}} = \mathcal{G}_{\mathrm{txt}}(o_i)$ in the same space as above. We then define the image-text similarity score for each candidate $i$,
\begin{equation}
\resizebox{0.45\linewidth}{!}{$%
S_i \;=\; \operatorname{sim}\!\big(\mathbf{e}_i^{\text{txt}},\,\mathcal{F}_D\big), 
\quad i = 1,\ldots,G
$}
\label{eq:vis-sim}
\end{equation}
where $\operatorname{sim}(\cdot,\cdot)$ denotes cosine similarity, and a larger $S_i$ indicates a stronger agreement between the report and disease-relevant visual features.

To obtain a scale-free visual signal, we convert these scores into rank-normalized visual rewards. 
Let $\mathrm{rank}(S_i)$ return $1$ for the largest similarity and $z$ for the smallest; we define,
\begin{equation}
r_i^{\mathrm{vis}}
\;=\;
1 \;-\; \frac{\mathrm{rank}(S_i)-1}{z-1},
\label{eq:vis-rank}
\end{equation}
so that $r_i^{\mathrm{vis}} \in [0,1]$ assigns the highest score to the most image-consistent candidate within each group.

We retain the scalar textual reward $R_{\text{text}}(o_i,g;k)$ from Eq.~\eqref{eq:stage1-text}, and then define the composite visual-alignment reward, 
\begin{equation}
\resizebox{0.65\columnwidth}{!}{$
R_{\text{vis}}(o_i,g,I;k)
=
(1 - \lambda_{\mathrm{vis}})\,R_{\text{text}}(o_i,g;k)
\;+\; \lambda_{\mathrm{vis}}\,r_i^{\mathrm{vis}}
$}
\label{eq:stage2-reward}
\end{equation}
where $\lambda_{\mathrm{vis}} \in [0,1]$ acts as the mixing coefficient that controls the strength of visual grounding.
The group-relative advantages are then obtained by rank-normalization over this composite reward,
\begin{equation}
\resizebox{0.5\linewidth}{!}{$%
A_i^{\mathrm{vis}} 
\;=\; \mathrm{RankNorm}\big(\{R_{\text{vis}}(o_j,g,I;k)\}_{j=1}^{G}\big).
$}
\label{eq:combined-adv}
\end{equation}
Thus, we simply replace Stage-1 advantages with $A_i^{\mathrm{vis}}$, which biases the policy toward candidates that are both textually faithful and visually aligned. Therefore, the optimization objective is given as,

\begin{equation}
\resizebox{0.85\columnwidth}{!}{$
\begin{aligned}
\mathcal{L}_{\mathrm{\ours{}}}(\theta)
&= -\,\frac{1}{G}\sum_{i=1}^{G}\frac{1}{|o_i|}\sum_{t=1}^{|o_i|}
\min\!\Big( r^{vis}_{i,t}(\theta)\,A^{vis}_{i,t}, \\
&\qquad\qquad\operatorname{clip}\!\big(r^{vis}_{i,t}(\theta),\,1-\epsilon,\,1+\epsilon\big)\,A^{vis}_{i,t} \Big)
\;-\; \beta\, D_{\mathrm{KL}}\!\left[\,\pi_{\theta}\,\middle\|\,\pi_{\mathrm{ref}}\,\right],
\end{aligned}
$}
\label{eq:grpo}
\end{equation}

\noindent where, $D_{\mathrm{KL}}\!\big(\pi_{\theta}\,\|\,\pi_{\mathrm{ref}}\big)$ serves as a regularization term penalizing divergence of the current policy $\pi_{\theta}$ from the reference policy $\pi_{\mathrm{ref}}$, with $\beta$ controlling the regularization strength w.r.t. the policy.

\noindent\underline{\textbf{Format Reward.}} To ensure that generated candidate reports follow the proper format, we impose a format reward (denoted as $R_\text{format}$) in both stages to check whether the generated report contains the required tag. Specifically, the report must be contained within the \texttt{<report>} and \texttt{</report>} tags,  respectively.

%% file: figures/pipeline.tex
\begin{figure*} [ht!]
    \centering
    \includegraphics[width=\linewidth]{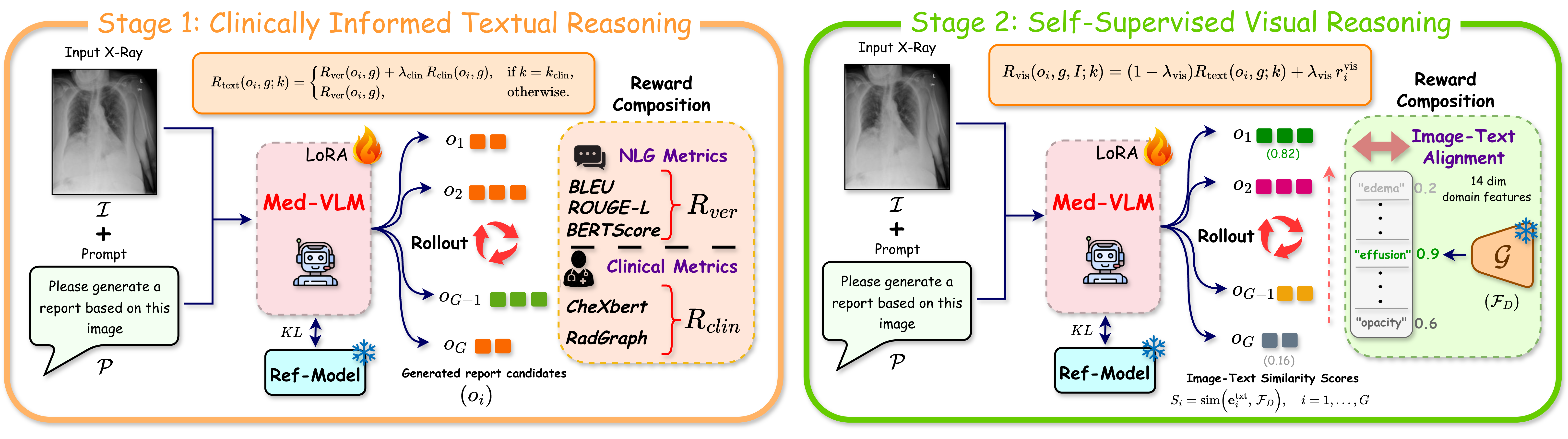}
    \caption{\textbf{Workflow of \ours{}.} \textbf{Left:} In Stage-1, \ours{} is trained with verifiable textual rewards $R_{ver}$ with periodic clinical guidance $R_{clin}$ every $k_{clin}$ policy steps to familiarize the base Med-VLM $\mathcal{M(\cdot)}$ with radiology reasoning patterns. \textbf{Right:} In Stage-2, \ours{} is further optimized with rewards aligned using \emph{image-text similarity scores} computed by using embeddings obtained utilizing the multi-label domain expert model $\mathcal{G}$, shifting the model's reasoning focus from textual knowledge, to grounded, region-wise disease reasoning capabilities in a self-supervised fashion. We also utilize \texttt{format} rewards to maintain structure of generated reports using \texttt{<report>} tags.}
    \label{fig:pipeline}
\end{figure*}

\vspace{-1em}

%% file: algo/algo.tex
\begin{algorithm}[t]
\caption{\textbf{\ours{}}: Clinically informative Multi-Stage Visual Alignment}
\label{alg:valor-final}
\setlength{\fboxsep}{3pt}
\colorbox{cvprblue!5}{%
\parbox{\dimexpr\columnwidth-2\fboxsep}{%
\begin{algorithmic}
\Require $\mathcal{D}=\{(\mathcal{I},g)\}$, Med-VLM $\mathcal{M}_\theta$, frozen expert $\mathcal{G}$, group size $G$
\State \textbf{Initialize:} $\theta \!\leftarrow\! \theta_{\mathrm{SFT}}$ (Eq.~\eqref{eq:sft}), $\theta_{\mathrm{ref}} \!\leftarrow\! \theta$, global step $k\!\leftarrow\!0$
\For{$t \in \{0,1\}$} \Comment{$t{=}0$: Stage-1 text alignment, $t{=}1$: Stage-2 visual alignment}
  \State Configure trainable modules of $\mathcal{M}_\theta$ for stage $t$
  \For{minibatch $\mathcal{B} \subset \mathcal{D}$}
    \State $k \leftarrow k+1$
    \For{each $(\mathcal{I},g)\in \mathcal{B}$}
      \State Sample $G$ candidate reports $\{o_i\}_{i=1}^{G}\sim \pi_\theta(\cdot\mid \mathcal{I},\mathcal{P})$
      \State Compute $R_{\mathrm{ver}}(o_i,g)$ (Eq.~\eqref{eq:ver-reward}) and $R_{\mathrm{clin}}(o_i,g)$ (Eq.~\eqref{eq:clin-reward})
      \State Form Stage-1 reward $R_{\mathrm{text}}(o_i,g;k)$ using periodic clinical injection (Eq.~\eqref{eq:stage1-text})
      \If{$t=1$}
        \State Compute image/text features and similarity $S_i$ (Eq.~\eqref{eq:vis-sim}); rank-normalize to $r_i^{\mathrm{vis}}$ (Eq.~\eqref{eq:vis-rank})
        \State Compose $R_{\mathrm{vis}}(o_i,g,\mathcal{I};k)$ (Eq.~\eqref{eq:stage2-reward}) and set $A_i \leftarrow A_i^{\mathrm{vis}}$ via Eq.~\eqref{eq:combined-adv}
      \Else
        \State Set $A_i \leftarrow \mathrm{RankNorm}\!\big(\{R_{\mathrm{text}}(o_j,g;k)\}_{j=1}^{G}\big)$ \Comment{Adv. computation}
      \EndIf
    \EndFor
    \State Update policy with KL regularization to $\pi_{\theta_{\mathrm{ref}}}$ (Eq.~\eqref{eq:grpo})
  \EndFor
\EndFor
\State \Return $\mathcal{M}_{\theta^\star}$
\end{algorithmic}
}}
\end{algorithm}

%% file: sec/06_experiments.tex
\section{Experiments}
\label{sec:expts}




\subsection{Experimental Settings} 
\textbf{Datasets and Evaluation Metrics.} We benchmark \ours{} on two widely used radiology report datasets: IU-XRay~\cite{demner2015preparing} and MIMIC-CXR~\cite{johnson2020mimic}. The generated reports are evaluated for quality across two critical categories: clinical accuracy and natural language generation quality. For clinical accuracy, we follow existing works~\cite{chen2020generating, chen2022cross, wang2023metransformer}, using 14 CheXpert~\cite{smit2020chexbert} labels to assess the clinical accuracy of the generated reports. Addtionally, we also evaluate on more clinically oriented metrics such as GREEN~\cite{ostmeier2024green} and RaTEScore~\cite{zhao2024ratescore}. In addition, we employ commonly used natural language metrics, such as BLEU Score~\cite{papineni2002bleu}, ROUGE~\cite{lin2004rouge} and METEOR~\cite{denkowski2011meteor} to assess the quality of the generated reports. 


\input{figures/results}

\noindent\textbf{Baselines and Implementation Details.} We compare against a set of direct baselines on top of the base Med-VLM including SFT, DPO~\cite{rafailov2023direct} and GRPO~\cite{shao2024deepseekmathpushinglimitsmathematical}. We also compare against recent post-alignment work MMedPO~\cite{zhu2024mmedpo} which uses Direct Preference Optimization (DPO), as well as GRPO without any visual alignment and clinical intervention, as outlined in Stage-1. We adopt LLaVa-Medv1.5~\cite{li2023llava} as our baseline Med-VLM, in line with \cite{zhu2024mmedpo} and provide additional results with QwenVL-2.5-7B~\cite{bai2025qwen3} in the supplementary. We also extensively compare our method on clinically informative using ground truth labels across 14 disease categories, evaluating F1 score, GREEN~\cite{ostmeier2024green} and RaTEScore~\cite{zhao2024ratescore}. Further, we compare \ours{} against zero-shot non-medical SOTA proprietary models such as GPT-4o~\cite{hurst2024gpt}, GPT-4.1~\cite{achiam2023gpt} Gemini-2.5 Pro~\cite{comanici2025gemini} and medical SOTA proprietary models MedFlamingo~\cite{moor2023med}, MAIRA-2~\cite{bannur2024maira} and HuatuoGPT-V-8B~\cite{chen2024towards} for generation quality. For the GRPO training phase, we utilize LoRA fine-tuning~\cite{hu2022lora} with a batch size of 8, a learning rate of \texttt{1e-6}, and perform the alignment operation for three epochs. Further details on the benchmarking, instruction prompts and additional implementation details are included in the supplementary.

\subsection{Comparison with State-of-the-Art Methods}

\noindent\textbf{Enhanced Clinical Reasoning and Generation Quality.} We highlight how \ours{} can significantly generate more visually aligned and clinically enriched reports in {\hyperref[fig:result]{Fig.~\ref{fig:result}}}, w.r.t. to the baseline Med-VLM and the ground truth report. Note that \ours{}, despite not using any externally trained model or curated expert preference data, is still able to capture the disease-relevant portions of the X-ray (top row) and is able to correctly capture the disease observations (such as ``\texttt{atelectasis}'', ``\texttt{cardiomegaly}'', ``\texttt{edema}'' etc.). In the second row, LLaVA-Med not only hallucinated disease observations that had no correlation with the input image but also generated generic text with no medical terminology showing suboptimal to no reasoning abilities at all. \ours{} without the visual alignment performs significantly better than the base Med-VLM but still fails in cases where the semantic knowledge of disease prevalent regions is needed. In contrast, \ours{} generates reports with disease-relevant observations that closely align with the input X-ray, demonstrating superior clinical reasoning abilities.

\input{figures/visualization}

\noindent\textbf{Visual Reasoning and Grounding Abilities.} We also visualize which portions of the image are being focused on by each baseline in {\hyperref[fig:attention]{Fig.~\ref{fig:attention}}} by visualizing the attention tokens of each model on the X-ray. Clearly, it can be seen that not only the baseline fails to focus on disease-relevant areas, but it also looks at regions that have no bearing on where the potential diseases can be, leading to inaccurate diagnoses, i.e. ``\texttt{no visible abnormalities}.'' \ours{} without the visual grounding proposed in Sec. \ref{subsec:rewards} performs relatively better and is able to detect the opacity in the right lung, but the greater focus on the left lung and pelvis region underlines the insufficient anatomical knowledge in the model. After performing visual alignment, the region where \ours{} focuses on becomes much more concentrated and accurate, which is reflected in the generated report which has the correct observations in this context ``\texttt{opacity}'', ``\texttt{atelectasis}'', and ``\texttt{consolidation}'', encouraging the model to \textit{look before generation}. \\

\input{tables/1_gen_clin_all}
\noindent\textbf{Comparison w.r.t. Post-Trained Models.} We evaluate \ours{} on two important tasks: generation quality and clinical accuracy on the two radiology report benchmarks IU-XRay~\cite{demner2015preparing} and MIMIC-CXR~\cite{johnson2020mimic} in {\hyperref[tab:gen_clinical_combined]{Table~\ref{tab:gen_clinical_combined}}}, outperforming state-of-the-art preference tuning method MMedPO~\cite{zhu2024mmedpo}, DPO~\cite{rafailov2023direct} and \ours{} without visual grounding in {\hyperref[tab:gen_clinical_combined]{Table~\ref{tab:gen_clinical_combined}}}. We obtain a significant improvement over MMedPO on all three natural language metrics and \ours{} with visual alignment. To assess clinical accuracy, we further report CheXpert-F1 accuracy and more clinically aligned metrics GREEN~\cite{ostmeier2024green} and RaTEScore~\cite{zhao2024ratescore} in {\hyperref[tab:clinical-acc]{\hyperref[tab:gen_clinical_combined]{Table~\ref{tab:gen_clinical_combined}}}. On IU-Xray, \ours{} achieves the best F1 score (\textbf{57.2}\%), clearly improving over LLaVA-Med (38.6\%) and MMedPO (49.8\%) by more than \textbf{~10\%}, on GREEN and RaTEScore by more than \textbf{~15\%} on average. As it can be seen from both benchmarks, the alignment step of our model is crucial for obtaining a large improvement. Overall, these results demonstrate that \ours{} not only produces more fluent and semantically aligned reports but also maintains strong clinical accuracy.

\noindent\textbf{Comparison w.r.t. SOTA Multimodal LLMs.} We also compare our model with mainstream proprietary multimodal LLMs such as GPT-4o~\cite{hurst2024gpt} and medical LLMs~\cite{moor2023med, bannur2024maira, chen2024huatuogpt} in a zero-shot setting in {\hyperref[tab:iu_zero_shot]{Table~\ref{tab:iu_zero_shot}}}. Our method outperforms GPT-4 series~\cite{hurst2024gpt}, Gemini~\cite{comanici2025gemini} by \textbf{~+10\%} and SOTA medical MLLMs by an average margin of \textbf{~+12\%} in generation quality. Therefore, \ours{} consistently performs better than both general-purpose and medical multimodal LLMs, showing more anatomy of visual knowledge along with enhanced clinical accuracy despite the significantly larger pre-training knowledge corpus of these models on large amounts of data.
\input{tables/03_zero_shot}

\subsection{Ablation Studies} 

\input{figures/clin_vis_abl}

\input{figures/nlg_vs_iteration}


\noindent\textbf{Need for Multi-Stage Reward Guidance.}
{\hyperref[fig:plots]{Fig.~\ref{fig:plots}(a)}} ablates the Stage-2 mixing weight $\lambda_{\mathrm{vis}}$ in Eq.~\eqref{eq:stage2-reward}. We observe that BLEU-4, ROUGE-L, and CheXbert F1 consistently peak when $\lambda_{\mathrm{vis}}\!\in\![0.4,0.6]$, indicating that neither purely textual nor purely visual optimization is sufficient. Lower weightage under-utilizes visual informativeness and reduces Stage-2 to Stage-1 behavior, whereas large values collapses training towards only disease terms. This confirms that gains of \ours{} arise from a balanced combination of textual faithfulness and visual grounding, which is further reflected in improvements reported in {\hyperref[tab:ablate_lvis]{Table~\ref{tab:ablate_lvis}}}.

\noindent\textbf{Clinical/Visual Intervention.}
{\hyperref[fig:plots]{Fig.~\ref{fig:plots}(b)}} studies the periodicity of clinical reward injection through $k_{\text{clin}}$. Very frequent supervision (small $k_{\text{clin}}$) destabilizes optimization and biases the policy toward short, overly conservative clinical phrases, while overly sparse supervision weakens the clinical signal and under-utilizes Stage-1 guidance. A moderate reward period yields the best trade-off, as also shown in {\hyperref[tab:ablate_kclin]{Table~\ref{tab:ablate_kclin}}}. {\hyperref[fig:nlg_vs_iter]{Fig.~\ref{fig:nlg_vs_iter}}} justifies this design by showing clear performance jumps both at the onset of periodic clinical guidance and again at the beginning of Stage-2, demonstrating that clinical intervention improves report faithfulness early on, while visual intervention further strengthens grounded generation. 
\input{tables/04_ablation}

\vspace{-1em}

%% file: figures/results.tex
\begin{figure*} [ht!]
    \centering
    \includegraphics[width=\linewidth]{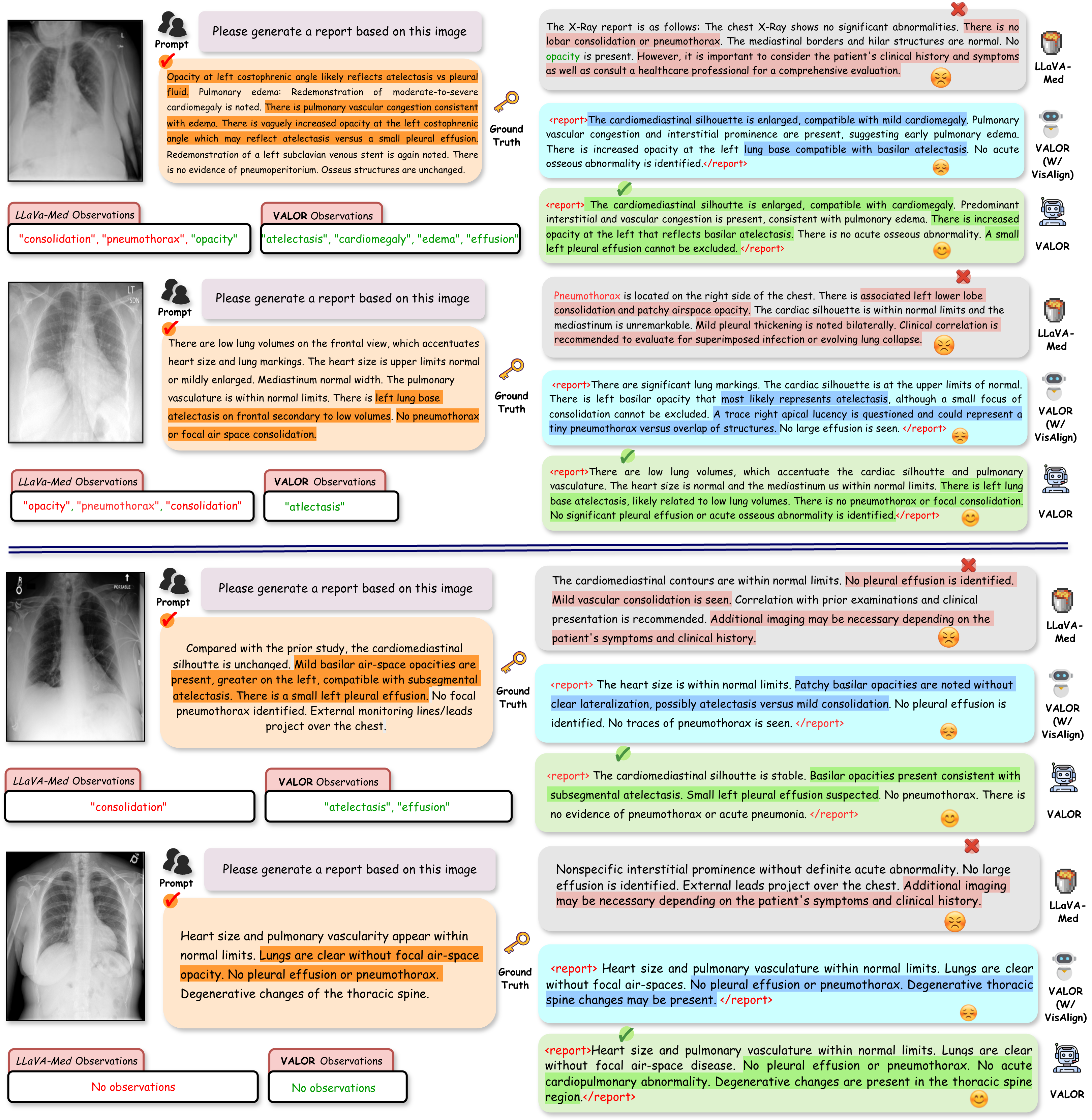}
    \caption{\textbf{Generated reports with \ours{}.} \textbf{First Case:} MedVLM~\cite{li2023llava} fails to detect subtle reasoning patterns needed for accurate diagnosis, resulting in hallucination of disease observations such as \texttt{``consolidation''}. In contrast, \ours{} with verifiable guidance leads to significantly better reasoning and diagnosis capabilities. With visual alignment, \ours{}'s enhanced multimodal reasoning leads to correct diagnosis of all observations,  showcasing the need for awareness of disease-relevant portions and focused attention. The MedVLM again hallucinates diseases due to insufficient visual focus, in comparison \ours{} reduces hallucinations and generates reports faithful to the X-Ray exhibiting better anatomical and semantic awareness. \textbf{Second Case:} Similarly in the second case, although MedVLM is able to deduce that the X-ray is benign, it’s reasoning is incorrect as highlighted in it’s generated report. VALOR comes to the same conclusion but demonstrating better overall reasoning capabilities. }
    \vspace{-1em}
    \label{fig:result}
\end{figure*}

%% file: figures/visualization.tex
\begin{figure*}[ht!]
    \centering
    \includegraphics[width=\linewidth]{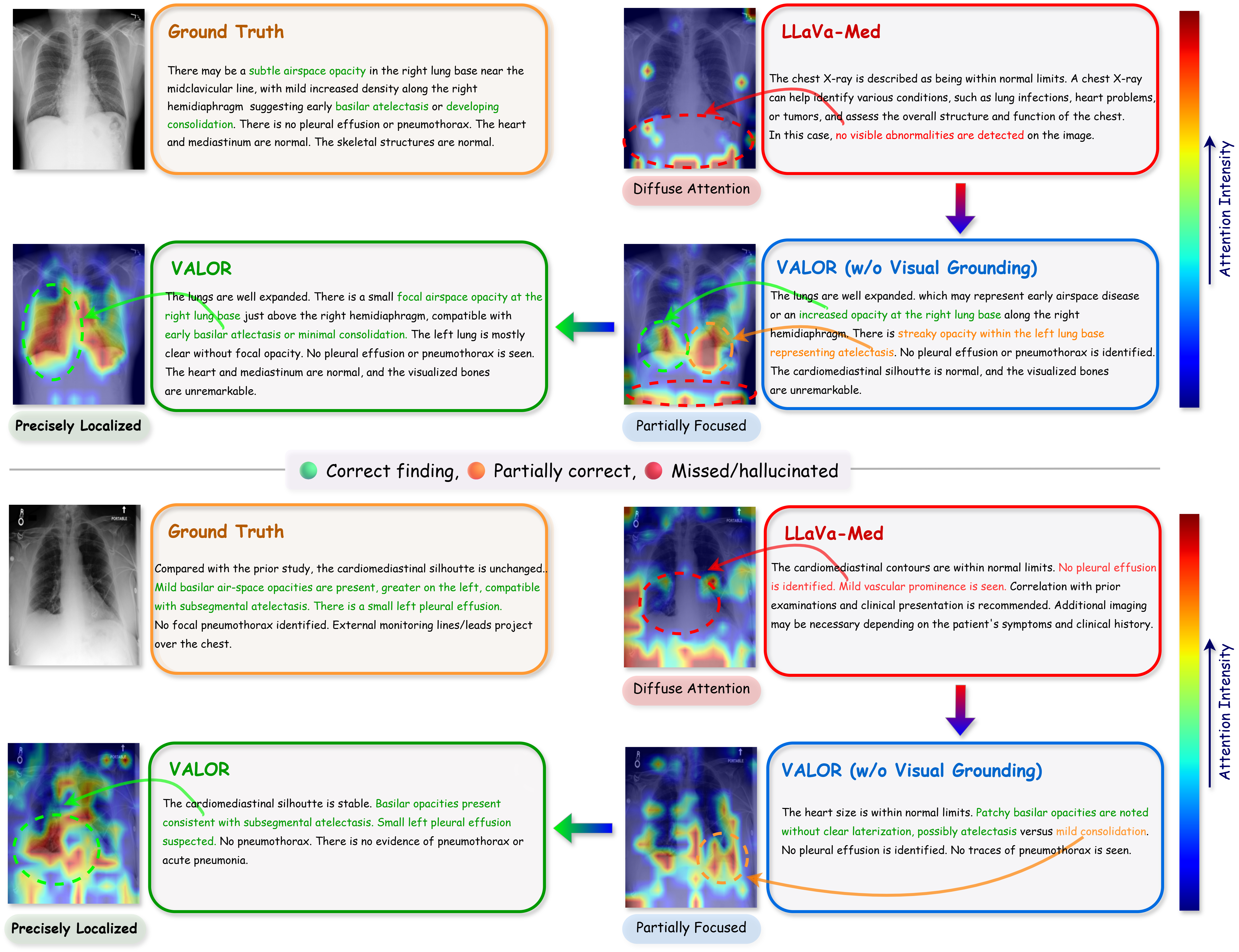}
    \caption{\textbf{Attention Map Visualization.} For the input X-Ray $\mathcal{I}$, we visualize the specific areas the Med-VLM~\cite{li2023llava} and \ours{} focuses on while generating the report. \textbf{Top Right, Both Cases:} is the attention map for the MedVLM~\cite{li2023llava} which misses the lung region due to lack of anatomical reasoning, leading to the generated report containing the inaccurate conclusion that no diseases can be detected. \textbf{Lower Right, Both Cases:} is the attention map without visual guidance results of \ours{}. It can be seen that it specifically learns to spot the clinically relevant regions for generating the report but still seems to focus on non-important regions at the base of the X-Ray. \textbf{Bottom Left, Both Cases:} In comparison, \ours{} allocates more attention on the right lung region, the most important area for this X-Ray demonstrating better visual localization and reasoning capabilities. \texttt{<report>} tags are omitted here for clarity.}
    \vspace{-2em}
    \label{fig:attention}
\end{figure*}

%% file: tables/1_gen_clin_all.tex
\definecolor{baselinebg}{HTML}{FFF8E1}   
\definecolor{advancebg}{HTML}{E3F2FD}    
\definecolor{oursrow}{HTML}{C8E6C9}      
\definecolor{ruleclr}{HTML}{BDBDBD}      
%

\newcommand{\NA}{\makebox[1.2em][c]{--}}
\newcommand{\NAb}{\cellcolor{baselinebg}\NA}
\newcommand{\NAa}{\cellcolor{advancebg}\NA}
\newcommand{\NAo}{\cellcolor{oursrow}\NA}

\newcommand{\grplabel}[2]{%
  \multirow{#1}{*}{\rotatebox[origin=c]{90}{\scriptsize #2}}}

\begin{table*}[t]
\centering

\caption{We compare the performance of \ours{} against the following baselines: base Med-VLM~\cite{li2023llava}, SFT, naive DPO~\cite{rafailov2023direct}, GRPO~\cite{shao2024deepseekmathpushinglimitsmathematical} and against SOTA post-trained model MMedPO~\cite{zhu2024mmedpo} on the generation quality~\cite{papineni2002bleu, lin2004rouge, denkowski2011meteor} of reports as well as clinical relevance and accuracy~\cite{smit2020chexbert, ostmeier2024green, zhao2024ratescore}. The effectiveness of visual alignment is seen in the last two rows in both benchmarks. Overall \ours{} achieves the best performance on both benchmarks. The best results are in \textbf{bold} and the second best in \textit{italics}.}
\label{tab:gen_clinical_combined}
\vspace{-1em}
\begingroup
\renewcommand{\arraystretch}{1.15}
\setlength{\tabcolsep}{4.5pt}
\arrayrulecolor{ruleclr}          
\resizebox{\textwidth}{!}{%
\begin{tabular}{
  p{3.2ex}          
  p{2.8ex}          
  l                  
  c c c              
  |                  
  c c c              
}

\arrayrulecolor{black}\toprule\arrayrulecolor{ruleclr}

& &
& \multicolumn{3}{c}{\textbf{\textit{Generation Quality}}}
& \multicolumn{3}{c}{\textbf{\textit{Clinical Accuracy}}} \\
\cmidrule(lr){4-6}\cmidrule(lr){7-9}

& &
  \multicolumn{1}{c}{\textbf{\textit{Model}}}
& \textbf{BLEU}~$\uparrow$
& \textbf{ROUGE-L}~$\uparrow$
& \textbf{METEOR}~$\uparrow$
& \textbf{F1}~$\uparrow$
& \textbf{GREEN}~$\uparrow$
& \textbf{RaTEScore}~$\uparrow$ \\

\arrayrulecolor{black}\midrule\arrayrulecolor{ruleclr}


\multirow{8}{*}[-1ex]{\rotatebox{90}{\textbf{IU-Xray}}}

& & LLaVA-Med v1.5
  & 8.55 & 7.58 & 4.56
  & 40.6 & 0.14 & 0.18 \\

\cmidrule(l){3-9}

& \grplabel{3}{Baselines}
& \cellcolor{baselinebg}+ SFT
  & \cellcolor{baselinebg}14.56 & \cellcolor{baselinebg}10.31 & \cellcolor{baselinebg}10.95
  & \cellcolor{baselinebg}38.8 & \cellcolor{baselinebg}0.11 & \cellcolor{baselinebg}0.16 \\

& & \cellcolor{baselinebg}+ DPO~\cite{rafailov2023direct}
  & \cellcolor{baselinebg}16.09 & \cellcolor{baselinebg}12.95 & \cellcolor{baselinebg}17.13
  & \cellcolor{baselinebg}40.5 & \cellcolor{baselinebg}0.12 & \cellcolor{baselinebg}0.16 \\

& & \cellcolor{baselinebg}+ GRPO~\cite{shao2024deepseekmathpushinglimitsmathematical}
  & \cellcolor{baselinebg} 17.90 & \cellcolor{baselinebg} 14.22 & \cellcolor{baselinebg} 19.60
  & \cellcolor{baselinebg}41.4 & \cellcolor{baselinebg}0.14 & \cellcolor{baselinebg}0.19 \\

\cmidrule(l){3-9}

& \grplabel{3}{Advancing}
& \cellcolor{advancebg}MMedPO~\cite{zhu2024mmedpo}
  & \cellcolor{advancebg}23.57 & \cellcolor{advancebg}29.52 & \cellcolor{advancebg}34.16
  & \cellcolor{advancebg}49.8 & \cellcolor{advancebg}0.26 & \cellcolor{advancebg}0.45 \\

& & \cellcolor{advancebg}\textbf{\ours} (w/o Visual Alignment)
  & \cellcolor{advancebg}25.61 & \cellcolor{advancebg}31.80 & \cellcolor{advancebg}35.90
  & \cellcolor{advancebg}55.8 & \cellcolor{advancebg}0.32 & \cellcolor{advancebg}0.52 \\

& & \cellcolor{oursrow}\textbf{\ours}
  & \cellcolor{oursrow}\textbf{27.03}
  & \cellcolor{oursrow}\textbf{33.10}
  & \cellcolor{oursrow}\textbf{36.88}
  & \cellcolor{oursrow}\textbf{57.2}
  & \cellcolor{oursrow}\textbf{0.34}
  & \cellcolor{oursrow}\textbf{0.53} \\

\arrayrulecolor{black}\midrule\arrayrulecolor{ruleclr}


\multirow{8}{*}[-1ex]{\rotatebox{90}{\textbf{MIMIC-CXR}}}

& & LLaVA-Med v1.5
  & 3.85 & 9.42 & 5.62
  & 50.2 & 0.08 & 0.11 \\

\cmidrule(l){3-9}

& \grplabel{3}{Baselines}
& \cellcolor{baselinebg}+ SFT
  & \cellcolor{baselinebg}11.11 & \cellcolor{baselinebg}9.38 & \cellcolor{baselinebg}7.71
  & \cellcolor{baselinebg}49.8 & \cellcolor{baselinebg}0.10 & \cellcolor{baselinebg}0.14 \\

& & \cellcolor{baselinebg}+ DPO~\cite{rafailov2023direct}
  & \cellcolor{baselinebg}11.20 & \cellcolor{baselinebg}9.45 & \cellcolor{baselinebg}7.81
  & \cellcolor{baselinebg}50.4 & \cellcolor{baselinebg}0.11 & \cellcolor{baselinebg}0.16 \\

& & \cellcolor{baselinebg}+ GRPO~\cite{shao2024deepseekmathpushinglimitsmathematical}
  & \cellcolor{baselinebg}12.26 & \cellcolor{baselinebg}10.58 & \cellcolor{baselinebg}9.22
  & \cellcolor{baselinebg}51.2 & \cellcolor{baselinebg}0.13 & \cellcolor{baselinebg}0.19 \\

\cmidrule(l){3-9}

& \grplabel{3}{Advancing}
& \cellcolor{advancebg}MMedPO~\cite{zhu2024mmedpo}
  & \cellcolor{advancebg}12.02 & \cellcolor{advancebg}\textbf{10.96} & \cellcolor{advancebg}\textit{9.88}
  & \cellcolor{advancebg}58.7 & \cellcolor{advancebg}0.22 & \cellcolor{advancebg}0.34\\

& & \cellcolor{advancebg}\textbf{\ours} (w/o Visual Alignment)
  & \cellcolor{advancebg}\textit{12.90} & \cellcolor{advancebg}10.30 & \cellcolor{advancebg}9.65
  & \cellcolor{advancebg}\textit{61.9} & \cellcolor{advancebg}\textit{0.26} & \cellcolor{advancebg}\textit{0.40} \\

& & \cellcolor{oursrow}\textbf{\ours}
  & \cellcolor{oursrow}\textbf{13.52}
  & \cellcolor{oursrow}\textit{10.85}
  & \cellcolor{oursrow}\textbf{9.95}
  & \cellcolor{oursrow}\textbf{63.7}
  & \cellcolor{oursrow}\textbf{0.29}
  & \cellcolor{oursrow}\textbf{0.46} \\

\arrayrulecolor{black}\bottomrule
\end{tabular}%
}
\vspace{-2em}
\endgroup
\end{table*}

%% file: tables/03_zero_shot.tex
\begin{table*}[t]
\centering
\caption{We compare the performance of \ours{} on both report generation benchmarks against proprietary SOTA medical and non-medical VLMs on generation quality of reports~\cite{papineni2002bleu, lin2004rouge, denkowski2011meteor}. \ours{} significantly outperforms in both settings highlighting the effectiveness of generating visually aligned reports. The best results are highlighted in \textbf{bold} and the second best in \textit{italics}.}
\label{tab:iu_zero_shot} 
\begingroup
\renewcommand{\arraystretch}{1.35}
\setlength{\tabcolsep}{6pt}
\resizebox{\textwidth}{!}{%
\begin{tabular}{l|l|ccc|ccc}
\toprule
\multirow{2}{*}{\textbf{Type}} & \multirow{2}{*}{\shortstack{\textbf{     Proprietary}\\\textbf{Models}}} &
\multicolumn{3}{c|}{\textbf{IU-Xray}} &
\multicolumn{3}{c}{\textbf{MIMIC-CXR}} \\
& & \textbf{BLEU} $\uparrow$ & \textbf{ROUGE-L} $\uparrow$ & \textbf{METEOR} $\uparrow$
  & \textbf{BLEU} $\uparrow$ & \textbf{ROUGE-L} $\uparrow$ & \textbf{METEOR} $\uparrow$ \\
\midrule

\multirow{2}{*}{General VLM}
  & GPT-4o~\cite{hurst2024gpt} 
  & 15.32 & 18.80 & 11.50
  & 12.10 & 17.20 & 10.30 \\
  & GPT-4.1~\cite{achiam2023gpt} 
  & 16.48 & 19.95 & 12.34
  & 12.94 & 17.88 & 10.92 \\
  & Gemini-2.5 Pro~\cite{comanici2025gemini} 
  & 17.26 & 21.10 & 13.05
  & 13.42 & 18.76 & 11.48 \\
  & \cellcolor{oursrow}\textbf{\ours}
  & \cellcolor{oursrow}\textbf{27.02} & \cellcolor{oursrow}\textbf{33.10} & \cellcolor{oursrow}\textbf{36.88}
  & \cellcolor{oursrow}\textbf{24.30} & \cellcolor{oursrow}\textbf{31.50} & \cellcolor{oursrow}\textbf{34.20} \\
\midrule

\multirow{2}{*}{Medical VLM}
  & MedFlamingo~\cite{moor2023med}
  & 14.85 & 24.10 & 22.44
  & 11.76 & 21.95 & 19.84 \\
  & MAIRA-2~\cite{bannur2024maira}
  & 20.22 & 30.60 & 32.36
  & 18.40 & 28.90 & 30.10 \\
  & HuatuoGPT-V-8B~\cite{chen2024huatuogpt}
  & 18.76 & 28.40 & 29.05
  & 17.12 & 27.35 & 28.44 \\
  & \cellcolor{oursrow}\textbf{\ours}
  & \cellcolor{oursrow}\textbf{27.02} & \cellcolor{oursrow}\textbf{33.10} & \cellcolor{oursrow}\textbf{36.88}
  & \cellcolor{oursrow}\textbf{24.30} & \cellcolor{oursrow}\textbf{31.50} & \cellcolor{oursrow}\textbf{34.20} \\
\bottomrule
\end{tabular}%
}
\endgroup
\vspace{-1em}
\end{table*}

%% file: figures/clin_vis_abl.tex
\begin{figure}
    \centering
    \vspace{-2em}
    \includegraphics[width=\linewidth]
    {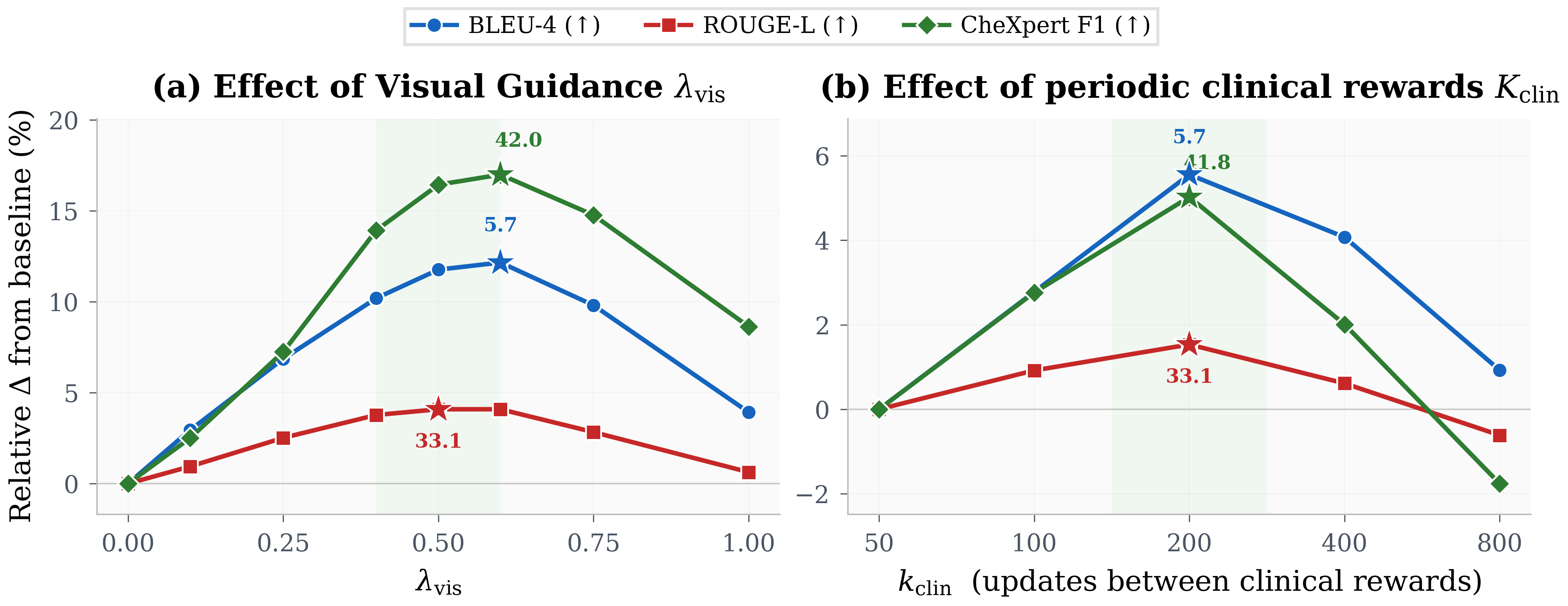}
    \caption{\textbf{Evolution of Clinical and visual periodic guidance.} \emph{Left:} \ours{} performs best with an optimal value of $\lambda_{\mathrm{vis}}\sim0.5$. \emph{Right:} \ours{} performs better with periodic injection of clinical rewards $R_{clin}$ per $k_{clin}$ steps instead of frequent (per-step) guidance which destabilizes performance.}
    \vspace{-2em}
    \label{fig:plots}
\end{figure}

%% file: figures/nlg_vs_iteration.tex
\begin{figure*} [ht!]
    \centering
    \includegraphics[width=\linewidth]{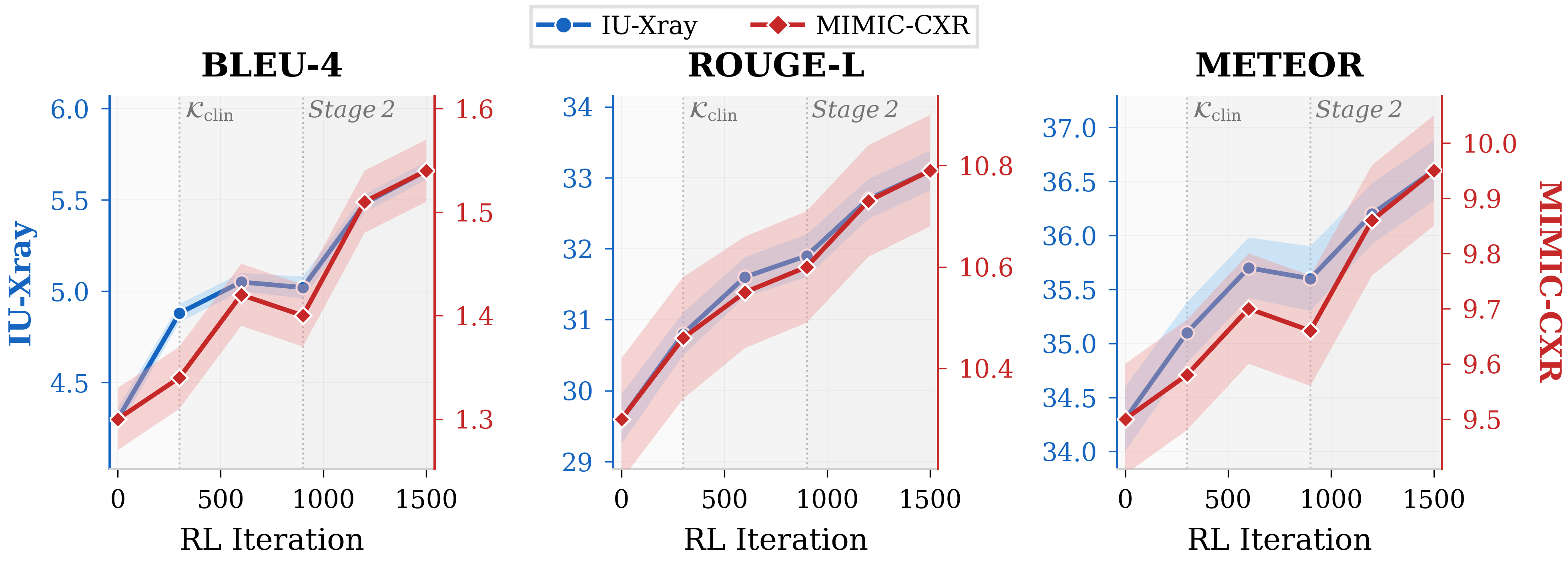}
    \caption{\textbf{Progression of generation quality with policy iterations.} We plot how the performance of the generation quality progresses with each policy iteration during Stage-1 and Stage-2. As it can be particularly seen in the \textit{leftmost} figure, there is a significant performance jump at the onset of the \textbf{clinical guidance} ($k_{clin}$) and at the \textbf{beginning of Stage-2} highlighting the importance of our two-stage multimodal alignment mechanism.}
    \label{fig:nlg_vs_iter}
\end{figure*}


%% file: tables/04_ablation.tex
\definecolor{collapsebg}{HTML}{FDEDED}   
\newcommand{\collapse}{\cellcolor{collapsebg}\textcolor{gray!70}{--\,\xmark}}

\begin{table}[t]
\centering
\setlength{\tabcolsep}{4pt}
\renewcommand{\arraystretch}{1.1}
\begin{minipage}[t]{0.48\linewidth}
\centering
\caption{\textbf{Stage-II visual sensitivity.}
Effect of $\lambda_{\mathrm{vis}}$ (Eq.~\eqref{eq:stage2-reward}) on BLEU and F1 Score.
\colorbox{collapsebg}{\collapse}~denotes non-convergence.}
\label{tab:ablate_lvis}
\resizebox{\linewidth}{!}{%
\begin{tabular}{c|cc|cc}
\toprule
\multirow{2}{*}{$\lambda_{\mathrm{vis}}$}
& \multicolumn{2}{c|}{\textbf{IU-Xray}}
& \multicolumn{2}{c}{\textbf{MIMIC-CXR}} \\
\cmidrule(lr){2-3}\cmidrule(lr){4-5}
& BLEU$\uparrow$ & F1$\uparrow$ & BLEU$\uparrow$ & F1$\uparrow$ \\
\midrule
0.0  & 25.61 & 55.8 & 12.90 & 61.9 \\
\rowcolor{oursrow}
\textbf{0.5}  & \textbf{27.03} & \textbf{57.2} & \textbf{13.52} & \textbf{63.7} \\
1.0  & \collapse & \collapse & \collapse & \collapse \\
\bottomrule
\end{tabular}}
\end{minipage}
\hfill
\begin{minipage}[t]{0.48\linewidth}
\centering
\caption{\textbf{Periodic clinical reward sensitivity.}
Effect of $k_{\mathrm{clin}}$ (Eq.~\eqref{eq:stage1-text}) on BLEU and F1 Score.
\colorbox{collapsebg}{\collapse}~denotes reward collapse during training.}
\label{tab:ablate_kclin}
\resizebox{\linewidth}{!}{%
\begin{tabular}{c|cc|cc}
\toprule
\multirow{2}{*}{$k_{\mathrm{clin}}$}
& \multicolumn{2}{c|}{\textbf{IU-Xray}}
& \multicolumn{2}{c}{\textbf{MIMIC-CXR}} \\
\cmidrule(lr){2-3}\cmidrule(lr){4-5}
& BLEU$\uparrow$ & F1$\uparrow$ & BLEU$\uparrow$ & F1$\uparrow$ \\
\midrule
dense & \collapse & \collapse & \collapse & \collapse \\
\rowcolor{oursrow}
\textbf{moderate} & \textbf{27.03} & \textbf{57.2} & \textbf{13.52} & \textbf{63.7} \\
rare & 26.30 & 55.1 & 12.14 & 62.4 \\
\bottomrule
\end{tabular}}
\end{minipage}
\vspace{-1.5em}
\end{table}

%% file: sec/07_conclusion.tex
\section{Conclusion}
\label{sec:conclusion}
\vspace{-0.5em}
We introduced \ours{}, a novel multimodal alignment framework that enhances the clinical and visual reasoning capabilities of existing Medical VLMs. To this end, we propose an alignment framework which utilizes both textual and visual rewards for optimization thereby enhancing the model's clinical multimodal comprehension capabilities. Our results demonstrate the versatility of \ours{} across a diverse set of datasets. Further our visualization studies show the effectiveness of our alignment approach, highlighting the crucial role the image plays in the overall task of radiology report generation. Our findings show \ours{}'s potential to advance medical image understanding and clinical healthcare. Future work can include more holistic reward designs and curriculum learning to clarify which components drive gains. These directions chart a practical path toward image-aligned and clinically reliable report generation systems.
